\documentclass[11pt]{article}

\usepackage[margin=1in]{geometry}
\usepackage[T1]{fontenc}
\usepackage[utf8]{inputenc}
\usepackage{newtxtext,newtxmath}
\usepackage{amsmath}
\usepackage{booktabs}
\usepackage{graphicx}
\usepackage{microtype}
\usepackage[hidelinks]{hyperref}

\newcommand{\hrr}{d_{\mathrm{hrr}}}
\newcommand{\bind}{\circledast}
\newcommand{\unbind}{\circledast^{\dagger}}

\title{Kathleen Remembers: Length-Invariant One-Shot Recall\\
Without Attention}
\author{George Fountzoulas\\
\small Department of Computer Engineering \& Informatics\\
\small Frederick University, Nicosia, Cyprus\\
\small \texttt{george.fountzoulas.research@gmail.com}}
\date{August 2026}

\begin{document}
\maketitle

\begin{abstract}
Recurrent, attention-free sequence models --- the family to which
Kathleen belongs, alongside state-space models and linear-attention
variants --- share a known structural weakness: a fading state cannot
perform exact recall of something seen once, far in the past. The
transformer answer, quadratic-cost attention over a growing window, is
exactly what this series set out to avoid. We add to the Kathleen
trunk a second memory layer --- a \textbf{notebook}: a fixed-key
holographic (HRR) associative store with a learned local write gate, a
self-gating raw read, and write-triggered forgetting --- 25K
parameters that attach to the logits of any trunk.
\textbf{(1) Mechanism.} On a controlled needle-in-haystack task, the
notebook reaches 80--82\% one-shot recall at $4\times$ the training
length (learned gates, two seeds) where the bare trunk scores
${\sim}4\%$ and a parameter-matched attention head scores 100\% inside
its training length and \textbf{0\% beyond it}. The route to this
result is reported as a ten-round, fully pre-registered forensic
sequence: three independent length leaks (gate receptive fields too
narrow to represent the correct write rule; $\ell_2$ normalization
amplifying empty-memory noise; trained trunk logits drifting out of
range at unseen lengths) are isolated with oracle gates,
counterfactual read-outs, and gate autopsies, and each is closed by
construction rather than by tuning. The final design's addressing is
length-invariant by construction --- fixed random content keys
composed over a 7-byte window with unitary role keys --- and the
memory alone, untrained, recalls at 90\% accuracy identically at 512,
2048 and 4096 bytes. Because the store is a linear superposition, two
capabilities follow from arithmetic alone: \textbf{selective
unlearning} --- one subtraction erases one fact to chance (4.3/3.1\%,
chance 3.8) while retained facts are unharmed (84\%), at both lengths
--- and \textbf{per-token attribution} --- counterfactual erasure
names the source fact of every correct byte with 100\% provenance,
at zero interference with ordinary prediction.
\textbf{(2) Real text.} On WikiText-2 bytes, the
notebook improves prediction of repeated rare words by $+0.15$--$0.27$
bits/byte over the same trunk, the gain \emph{growing} with the
distance between mentions and holding zero-shot at $4\times$ training
length, at zero overall cost; the write gate learns with no
supervision to spend its ink on content words ($\beta = 0.28$ on
repeated-word bytes vs 0.19 elsewhere). Write-triggered forgetting ---
decay tied to write mass, not to time, so no clock re-enters the
design --- eliminates the only observed failure mode (memory pollution
at $8\times$ length: first-mention cost $+0.33 \to -0.004$).
\textbf{(3) Scope and scale.} Two honest boundaries: a
parameter-matched attention head \emph{does} generalize on
natural-text repetition (its collapse is specific to surgical one-shot
recall), so the notebook's claim is exact recall at $O(L)$, not
repetition in general; and attached to a word-level model the notebook
is largely absorbed --- exact recall needs questions with exact
answers, which bytes provide and words do not. Within its habitat the
value \emph{scales}: on a WikiText-103 ladder ($8 \to 32 \to 128 \to
512$\,MB) the notebook's zero-shot repeat gain rises monotonically
($+0.06 \to +0.11 \to +0.15 \to +0.16$ bits/byte, two seeds at the top
rung) at zero-to-negative overall cost throughout. Memory, in this
family, is not a small-model crutch but a capability that matures with
the model. All experiments are pre-registered, seeds reported, and
reproducible on a single free-tier GPU.
\end{abstract}

\section{Introduction}

Papers 1--3 of this series built a byte-level, attention-free
architecture --- wavetable encoding, multi-scale reverberant state,
causal convolutions --- and showed it classifies~\cite{kathleen2},
learns without pretraining, and writes~\cite{kathleen3}, at
sub-million parameter scale, beating parameter-matched transformers on
data scaling throughout. Paper 3 closed with a diagnosis rather than a
victory lap: the reverberant state is a \emph{lossy compressor by
design}. It holds a fading summary of everything, and therefore an
exact copy of nothing. Ask a Kathleen model to reproduce a key--value
pair it saw once, two thousand bytes ago, and it fails --- not from
lack of scale but from the shape of its memory. The same diagnosis
applies, in the published record, to the whole attention-free family:
state-space models~\cite{gu2022s4,gu2023mamba} and linear-attention
variants~\cite{katharopoulos2020,peng2023rwkv} trade the transformer's
total-recall window for $O(L)$ cost, and give up exact long-range
recall in the bargain.

This paper adds the missing organ. The design goal was set by three
constraints the series already committed to: \textbf{$O(L)$} (no
quadratic window), \textbf{constant state} (a fixed-size memory, not a
growing cache), and \textbf{byte-native} (no tokenizer, no positional
table). The result --- we call it the \textbf{notebook} --- is a
25K-parameter module that attaches to the logits of any trunk:

\begin{itemize}
\item \textbf{fixed content keys}: every byte value owns a random unit
  vector in $d=2048$ dimensions, drawn once and never trained; keys
  for a position are composed from the last 7 bytes with unitary role
  vectors (an $n$-gram bundle in the HRR/holographic
  tradition~\cite{plate1995hrr,kanerva2009}). Addresses depend on
  \emph{what} the text says, never on \emph{where} --- length
  generalization by construction, not by hope.
\item \textbf{a learned local write gate}: a small causal convolution
  over a 16-byte window decides what is worth writing. It cannot see
  sequence length, absolute position, or global state --- so it cannot
  learn a length-dependent policy even by accident.
\item \textbf{a self-gating read}: the retrieved vector's raw
  magnitude is the gate. An empty or irrelevant memory returns
  near-zero and adds nothing; a hit returns a large vector and speaks.
  No read gate is learned, because none is needed.
\item \textbf{write-triggered forgetting}: each write erases a little
  of what came before ($\mathrm{mem} \leftarrow
  (1-g\beta)\,\mathrm{mem} + \beta\,\mathrm{write}$). Decay is tied to
  write mass --- to content --- not to elapsed time, so no clock
  re-enters the design and junk cannot accumulate with length.
\end{itemize}

The contributions, in the order the paper argues them:

\begin{enumerate}
\item \textbf{A mechanism result.} One-shot key--value recall at
  $4\times$ the training length with fully learned gates (80--82\%,
  two seeds), where the bare recurrent trunk scores ${\sim}4\%$ and a
  parameter-matched attention head scores 0\% (Section~3). The memory
  path alone, untrained, is length-flat to at least $8\times$ (90\% at
  512/2048/4096).
\item \textbf{A forensic method result.} The path to (1) required
  isolating three independent length leaks --- none of which was the
  memory itself. We argue the isolation tools (oracle gates,
  counterfactual read-outs, gate autopsies at two lengths,
  pre-registered verdicts for every round) are as reusable as the
  design (Section~3.3).
\item \textbf{An algebraic-consequences result.} Because the store is
  a linear superposition, erasure is subtraction and provenance is
  counterfactual subtraction: one fact is unlearned to chance with the
  others unharmed, and every correct byte names its source fact with
  100\% provenance --- both length-invariant, both without retraining
  (Section~4).
\item \textbf{A real-text result.} The notebook pays on natural text
  exactly where theory predicts --- repeated rare words at distance
  --- at zero overall cost, with an unsupervised write gate that
  discovers content-word salience (Section~5).
\item \textbf{Two honest boundaries.} Attention does generalize on
  loose natural-text repetition (its zero-shot collapse is specific to
  exact recall); and the notebook is absorbed by a word-level model,
  because words do not ask questions with exact answers (Section~6).
\item \textbf{A scaling result.} On an $8 \to 512$\,MB data ladder the
  notebook's zero-shot value rises monotonically ($+0.06 \to +0.16$
  bits/byte, two seeds at the top rung, zero-to-negative overall cost
  throughout): memory in this family matures with the model instead of
  being replaced by it (Section~7).
\end{enumerate}

\section{The two-layer memory}

\subsection{The trunk (recap)}

The Kathleen trunk is unchanged from Papers 2--3: byte embedding, $N$
blocks of \{multi-scale reverb bank $\|$ causal depthwise convolution,
sigmoid-mixed, $+$ FFN\}, layer norm, and a linear head. The reverb
bank is a bank of leaky integrators with input-dependent decay in
three half-life regimes (fast/medium/slow) --- the ``timing layer'':
it knows \emph{that} something happened and roughly when, but stores a
fading mixture, not retrievable items. All models in this paper use
$d = 96$--$128$, 2--3 blocks, 0.2--0.5M parameters.

\subsection{The notebook}

One associative store per model, holographic (circular-convolution
binding in FFT space~\cite{plate1995hrr}), dimension $\hrr = 2048$.
Write, at every position $t$:
\begin{align}
\mathrm{key}_t &= \ell_2\Big( \textstyle\sum_{j=0}^{6}
  \mathrm{role}_j \bind \mathrm{ck}[x_{t-j}] \Big)
  &&\text{(fixed)} \\
\beta_t &= \sigma\big(\mathrm{conv}_{16}(x)_t\big)
  &&\text{(learned)} \\
\mathrm{mem} &\leftarrow (1 - g\,\beta_t)\,\mathrm{mem} +
  \beta_t\,\big(\mathrm{key}_{t-1} \bind \mathrm{ck}[x_t]\big) &&
\end{align}
Read, at every position $t$:
\begin{align}
r_t &= \mathrm{mem} \unbind \mathrm{key}_t &&\text{(unbind)} \\
\mathrm{logits} &\mathrel{+}= s \cdot \big(r_t \cdot
  \mathrm{CK}^{\top}\big) &&\text{(raw)}
\end{align}
with $\bind$ circular convolution, $\unbind$ correlation (unbinding),
$\mathrm{ck}[b]$ the fixed random unit key of byte $b$, $\mathrm{CK}$
the $256 \times \hrr$ key table, $\mathrm{role}_j$ fixed unitary
vectors (unit magnitude spectrum, random phase), $g$ a learned forget
rate ($\sigma(\cdot)\cdot 0.2$), and $s$ a learned scalar. Cost per
position is $O(\hrr \log \hrr)$; the whole scan is parallel
(cumulative sums, chunked where forgetting is active). Parameter
count: the conv gate $+$ two scalars $\approx$ 25K; the key tables are
buffers, not parameters.

\subsection{Why each piece is what it is}

Every design choice above is the survivor of a measured failure,
reported in Section~3:

\begin{itemize}
\item \emph{Fixed keys, not learned projections of the hidden state}:
  learned state-keys inherit the state's length-dependence; fixed
  content keys are length-blind by construction (Round~6, after
  Rounds~1--5 established the delta-rule/learned-key variant fails
  zero-shot).
\item \emph{Local write gate, not a gate from the hidden state}: the
  trunk's hidden state drifts at unseen lengths, and a gate computed
  from it sags measurably ($\beta$ at needles: 0.49 at $L=512$ $\to$
  0.11 at $L=2048$, Round~8 autopsy). A byte-local gate cannot sag
  with length. Its window must be $\geq$ the needle span: with an
  8-byte window the gate \emph{cannot represent} the correct write
  rule and never trains (Rounds~7--8, diagnosed in Round~9) --- the
  fix is width, not optimization.
\item \emph{Raw self-gating read, not $\ell_2$-normalized read with a
  learned gate}: normalizing the retrieved vector amplifies
  empty-memory noise to unit norm at every position; training then
  shrinks the output scale, muting real hits along with the noise
  (Round~8's oracle arm trained to only 55--65\% for exactly this
  reason). Raw magnitudes carry the hit/no-hit information for free.
\item \emph{Write-triggered forgetting, not time decay and not
  delta-rule erasure}: time decay would reintroduce a clock (and
  length dependence); and with unitary keys the delta rule's targeted
  erasure is \emph{algebraically equivalent} to global decay --- a
  unitary key's power spectrum is flat, so ``erase along this key''
  and ``erase everything a little'' are the same operation
  (Section~5.3). The honest option is the cheap one.
\item \emph{Variable-length training (final ingredient, in the trunk
  not the notebook)}: a trunk trained at one fixed length learns
  readout weights tuned to slow-integrator values that sit elsewhere
  at $4\times$ length; its logits drift and shout over a perfectly
  correct memory (Round~9/10: memory-only counterfactual 90\% at 2048
  while the full model scored 0.5\%). Training on lengths
  $\{128 \dots 512\}$ cures the trunk; $2048+$ remains genuinely
  zero-shot.
\end{itemize}

\section{The needle laboratory}

\subsection{Task}

Byte-level haystack of random filler words. One or more needles ---
\texttt{<MARK> key value} ($6+6$ bytes) --- planted at uniform
positions; a query \texttt{<MARK> key ?} at the end; the model must
emit \texttt{value} byte-by-byte, teacher-forced accuracy measured on
the value bytes, binned by needle$\to$query distance. Train at
$L=512$ (later: variable 128--512), evaluate in-distribution and
zero-shot at $L=2048$ (and 4096 in diagnostics). Multi-needle variants
plant 4 pairs (interference); no-mark variants remove \texttt{<MARK>}
(salience).

\subsection{Headline arms (final round)}

\begin{figure}[t]
\centering
\includegraphics[width=0.95\textwidth]{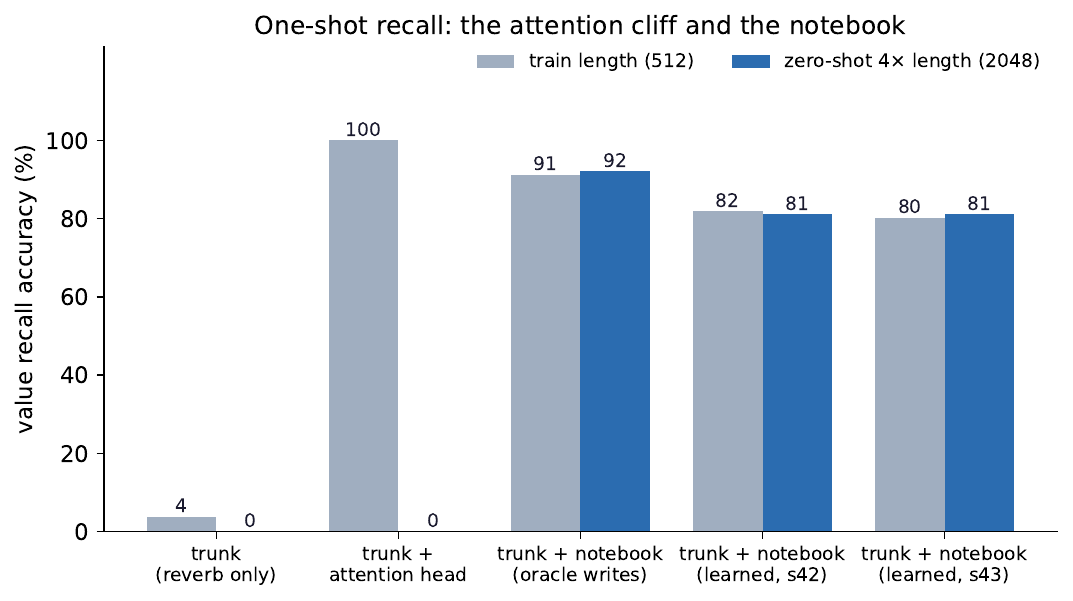}
\caption{One-shot recall at training length (512) and zero-shot at
$4\times$ length (2048). The attention head is perfect inside its
training length and zero beyond it; the notebook is length-invariant.}
\label{fig:needle}
\end{figure}

Identical trunks, identical data, identical budget; two seeds for the
learned arm (Figure~\ref{fig:needle}):

\begin{table}[h]
\centering
\small
\begin{tabular}{lccc}
\toprule
arm & params & train (far bin, $L=512$) & zero-shot $L=2048$ (far bin) \\
\midrule
trunk (reverb) only & 195K & 3.8\% & 0.0\% \\
trunk + attention head & 211K & \textbf{100\%} & 0.0\% \\
trunk + notebook (oracle write mask) & 195K & 91.2\% & \textbf{92.1\%} \\
trunk + notebook (learned local gates, s42) & 212K & 81.9\% & \textbf{81.1\%} \\
trunk + notebook (learned local gates, s43) & 212K & 80.2\% & \textbf{81.1\%} \\
\bottomrule
\end{tabular}
\caption{Needle-in-haystack recall, final round. All arms share
trunk, data, and budget.}
\label{tab:needle}
\end{table}

The counterfactual read-out (same trained weights, logits from one
component at a time) attributes the recall entirely to the memory:
memory-only $\approx$ full model at every length; trunk-only $\approx$
chance. The attention head is the honest mirror: perfect inside its
training length, zero beyond it --- the known zero-shot length cliff
of softmax attention~\cite{press2022alibi,peng2024yarn}, reproduced at
200K scale.

\subsection{The forensic path (ten rounds, all pre-registered)}

We report the route because the failures carry the information:

\begin{itemize}
\item \textbf{R1 --- proof of life.} Delta-rule fast-weight
  notebook~\cite{schlag2021} with gates learned from the hidden state:
  100\% train, 97\% zero-shot. First evidence the two-layer design can
  work at all.
\item \textbf{R2 --- it breaks.} Same design, 4 needles or no markers:
  total failure (3--4\%), and zero-shot at $4\times$ collapses to 0\%
  in every multi-needle configuration tested thereafter.
\item \textbf{R3--R5 --- discovery dynamics, and a harness bug.}
  Longer budgets and curricula: multi-needle \emph{is} learnable (a
  discrete ``click'' arrives after 2--6K steps, timing
  seed-dependent); a construction-order confound (modules built before
  seeding) was found and fixed --- reported because silent RNG-order
  bugs of this kind can flip qualitative conclusions and rarely get
  documented.
\item \textbf{R6 --- the archive lesson.} Replacing learned state-keys
  with fixed content keys (from this group's earlier HRR work) removes
  the learning cliff entirely (68\% at step 500 vs thousands of dark
  steps), half-solves salience (no-mark: 4\% $\to$ 53--64\%), and
  exposes a capacity ceiling (${\sim}80\%$ at $d=512$ with
  indiscriminate writing). Zero-shot: still 0 --- the keys were never
  the leak.
\item \textbf{R7 --- capacity vs gates, $2\times2$.} $d=2048$ lifts
  training to 96--97\% (noise was the plateau); zero-shot still
  ${\sim}0$; gates computed from a local byte window fail to train at
  all (3--4\%).
\item \textbf{R8 --- oracle round.} Perfect byte-computed write gates,
  no learning in the gate path: zero-shot \textbf{still 0} --- the
  verdict says the leak lives in the memory path. The gate autopsy
  simultaneously documents the trunk-gate sag ($\beta$@needle
  $0.49 \to 0.11$ at $4\times$). Two leaks visible at once; neither is
  the memory.
\item \textbf{Local diagnostics (CPU, zero training).} The memory
  alone --- oracle writes, no trunk, no gradient --- scores 58--68\%
  \emph{flat} at 512/2048/4096; with a clean write mask,
  \textbf{90.3/89.6\%} at 512/2048. Twenty seconds on a laptop settles
  what three GPU rounds circled: the memory is length-proof;
  everything else leaks. The $\ell_2$ noise-amplification and the
  8-byte gate myopia fall out of the same session.
\item \textbf{R9 --- assembly.} Raw self-gating read $+$ 16-byte local
  gate: local gates train for the first time (85--86\%, two seeds; the
  failure was representational width, not optimization). Zero-shot
  0.5\% --- by elimination (the memory path has one trained scalar),
  the \emph{trained trunk logits} drift at unseen lengths. The
  untrained trunk is verified length-stable; training creates the
  drift.
\item \textbf{R10 --- the cure.} Variable-length training (128--512).
  Oracle: $91 \to 92\%$ (no degradation at $4\times$); learned gates:
  $80$--$82\% \to 81\%$, two seeds, flat across distance bins.
  Verdict: the synthetic arc is complete.
\end{itemize}

Three separate leaks --- gate representational width, read-path noise
amplification, trunk logit drift --- each invisible while the others
were present, each isolated by an oracle or counterfactual rather than
by tuning. None of them was the associative memory, which was
length-invariant from the day it had content keys.

\section{Consequences of the algebra: unlearning and attribution}

Because the store is a linear superposition of key--value bindings,
two capabilities follow from the arithmetic itself --- no retraining,
no new parameters, no architectural change. We validate both on the
multi-needle rig (5 facts per window, $\hrr = 2048$, learned write
gates, trained as in Section~3), at the training length and zero-shot
at $4\times$.

\subsection{Selective unlearning by subtraction}

To erase one fact, subtract its writes: recompute the bound vectors
of the fact's span from the input (the keys are fixed and
content-derived, so this needs no stored bookkeeping beyond the span
itself) and subtract them from the memory vector. One vector
subtraction, cost $O(\text{span})$; the other facts are untouched.

\begin{table}[h]
\centering
\small
\begin{tabular}{lcccc}
\toprule
 & before & erased fact & retained facts & memory-only, erased \\
\midrule
$L=512$ (train) & 82.5\% & \textbf{4.3\%} & 84.1\% & 3.5\% \\
$L=2048$ (zero-shot) & 81.0\% & \textbf{3.1\%} & 84.5\% & 2.6\% \\
\bottomrule
\end{tabular}
\caption{Algebraic unlearning (chance $= 3.8\%$). The erased fact
drops to chance; the four retained facts are unharmed; the
memory-only counterfactual confirms the fact is \emph{absent from the
store}, not merely outvoted.}
\label{tab:unlearn}
\end{table}

The erased fact falls to chance level while the retained facts are,
if anything, marginally cleaner (less superposition crosstalk), and
the effect is length-invariant (Table~\ref{tab:unlearn}). The
memory-only read-out at the erased fact's query is itself at chance:
the information is gone from the store, not suppressed downstream.
This is forgetting with a guarantee --- the mechanism-level primitive
that data-deletion obligations (e.g.\ GDPR's right to erasure) ask of
deployed models and that gradient-trained weights cannot offer without
retraining.

\subsection{Per-token attribution}

The same subtraction, used counterfactually, yields provenance: for
each emitted byte, erase each stored fact in turn and measure the drop
in the chosen byte's logit; the fact whose erasure drops it most is
the primary source. Component attribution (trunk vs.\ notebook) falls
out of the two-path logit sum for free.

At both lengths, on bytes the model gets right: \textbf{provenance is
100\%} --- every correct value byte names the fact (and hence the
write position) it came from; 96.4/95.2\% of correct value bytes are
memory-decisive (the trunk alone would not have produced them); the
notebook's interference with ordinary filler prediction is
$+0.15$/$+0.03$ percentage points (i.e.\ none). The self-gating
margin separates hit from no-hit reads by $4.1\times$/$3.9\times$
--- 93\%/87\% of the measured oracle ceiling ($4.4\times$), which is
set by HRR superposition crosstalk, not by the gate. The model can
therefore emit, alongside every byte, a faithful audit line of the
form \emph{``notebook, fact \#2, written at position 255, logit drop
1.1''} --- a property attention heat-maps approximate but do not
guarantee~\cite{wu2022memorizing}, and one we exploit again in the
unlearning demo above: attribution and erasure are the same
subtraction read in two directions.

\section{Real text: the repeat probe}

\subsection{Probe}

No marks, no planted needles. On WikiText-2
bytes~\cite{merity2017wikitext}, partition test-window word bytes: a
word (alphabetic, $\geq 4$ chars) whose same form occurred $\geq 64$
bytes earlier in the window is a \textbf{repeat}; first occurrences
are the control. Report bits/byte on each set, and \textbf{extra
repeat gain}: ($\mathrm{bpb}_{\mathrm{first}} -
\mathrm{bpb}_{\mathrm{repeat}}$) minus the same difference for the
bare trunk --- what the notebook adds beyond the trunk's own handling
of repetition. Train varlen 128--512; evaluate at 512 and zero-shot
2048/4096.

\subsection{Result}

\begin{figure}[t]
\centering
\includegraphics[width=0.7\textwidth]{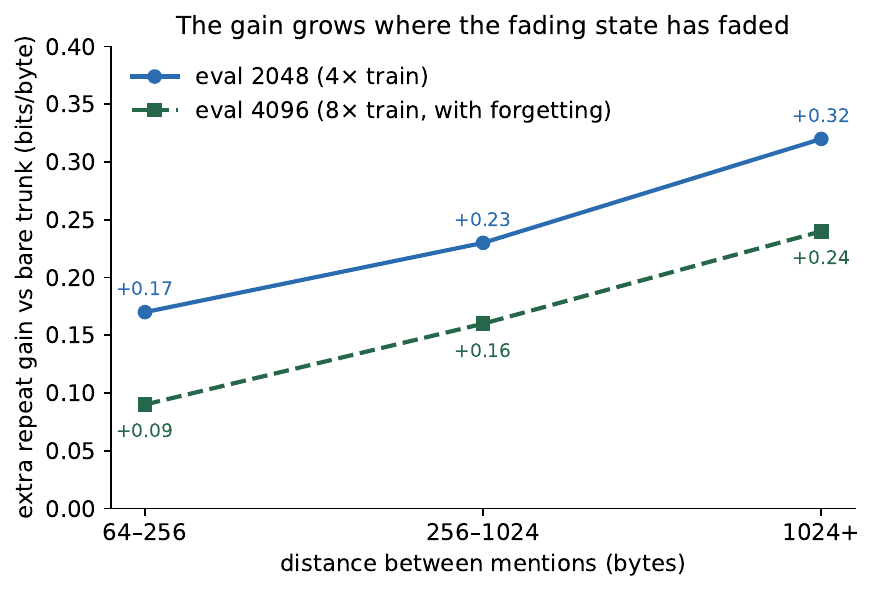}
\caption{Extra repeat gain vs distance between mentions (WikiText-2
bytes). The notebook's benefit grows exactly where the fading state
has faded, and holds at $4\times$ and $8\times$ the training length.}
\label{fig:distance}
\end{figure}

Two seeds: extra repeat gain \textbf{$+0.19$/$+0.15$ at 512,
$+0.27$/$+0.24$ at 2048} --- the gain \emph{grows} where the reverb
state has faded (Figure~\ref{fig:distance}), which is the two-layer
design's signature prediction. Overall bpb cost: $-0.02$ (the notebook
arm is marginally better overall). The write gate, trained only by the
language-modeling loss, discovers salience: $\beta = 0.28$ on
repeated-word bytes vs 0.19 elsewhere. The bare trunk's own repeat
``gain'' is negative ($-0.16$): repeated rare words are \emph{harder}
than first mentions for a fading state, not easier.

\subsection{The one failure, and forgetting}

At 4096 ($8\times$ training length) the plain notebook begins to
charge rent: first-mention bpb degrades $+0.33$ over the trunk ---
with $\beta \approx 0.19$ everywhere, thousands of positions fill the
store with junk and superposition noise leaks into ordinary
prediction. Write-triggered forgetting ($\mathrm{mem} \leftarrow
(1-g\beta)\,\mathrm{mem} + \beta\cdot\mathrm{write}$, $g$ learned)
eliminates the cost entirely (\textbf{$+0.33 \to -0.004$}) while
keeping every distance bin's gain positive ($+0.09$/$+0.16$/$+0.24$ at
4096). The trade is real --- near-bin gains shrink (the eraser takes
some useful ink too) --- and it is near-optimal for this mechanism:
with unitary keys, delta-rule targeted erasure reduces algebraically
to global decay (flat key spectrum $\Rightarrow$ ``erase this key''
$\equiv$ ``erase everything slightly''), so no smarter eraser exists
inside the FFT superposition design.

\subsection{Scope: what attention does here}

A parameter-matched one-head attention twin, same trunk, same
training: on \emph{natural} text its repeat gains are large
($+0.83 \dots +1.04$) and --- unlike on the needle --- hold at 2048
zero-shot. Loose copying over many mentions and soft context matches
is attention's home game, and it generalizes there. The needle
collapse is specific to surgical one-shot recall. The claim this paper
makes is therefore scoped precisely: \textbf{exact recall of once-seen
material at $O(L)$ and constant state} --- plus the practical note
that the attention twin could not even be \emph{evaluated} at 4096
within the same memory budget (its $O(L^2)$ matrices do not fit),
while the notebook runs at any length.

\section{The wrong habitat: a word-level negative result}

Attached to the series' word-level generation model (Paper 3's
composer: 10K vocabulary, ${\sim}3$M params), the same recipe ---
content keys per word, 3-gram bundles, 8-word gate --- is largely
absorbed: extra repeat gain $+0.07$/$-0.06$ at the training length
(two seeds), $+0.08$/$+0.13$ at $4\times$ zero-shot, gate selectivity
none ($\beta$ flat at 0.16). Cost, as everywhere: zero.

The explanation is structural, and we consider it a finding rather
than a disappointment. Exact-recall memory answers questions that have
exact answers. At byte level such questions are everywhere: having
seen \texttt{wond}, the continuation \texttt{erful} is
\emph{deterministic given the memory}. At word level the question
``which word follows?'' almost never has an exact answer --- the
surrounding words of a repeated mention differ from its first
occurrence, so a key built from context matches nothing. The
notebook's habitat is bytes and signals: exactly the levels where this
series operates and where tokenized models do not.

\section{Scaling: the ladder}

\begin{figure}[t]
\centering
\includegraphics[width=0.7\textwidth]{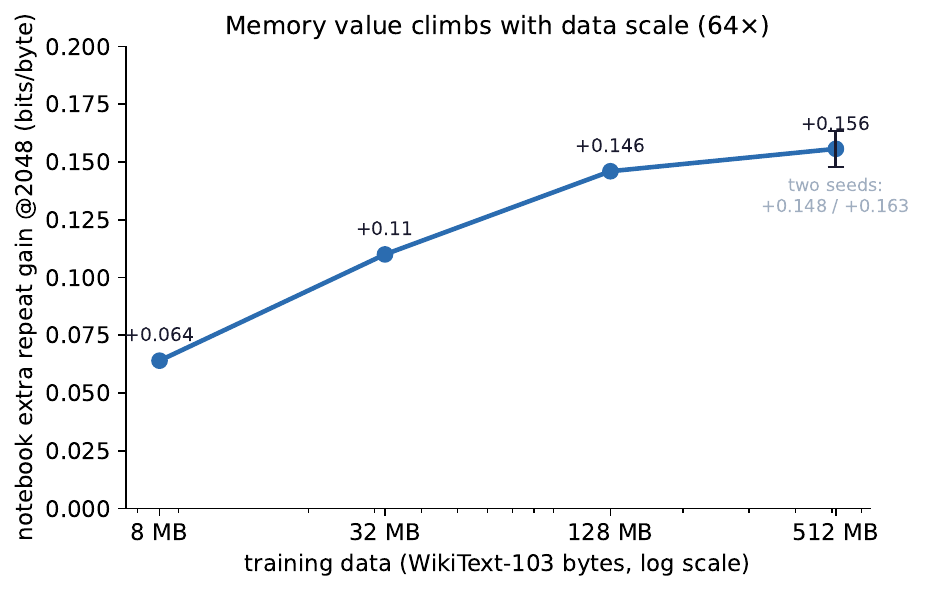}
\caption{Notebook extra repeat gain at $8\times$ training length vs
training data ($\log$ axis). $64\times$ data, $2.4\times$ gain;
two-seed error bar at the top rung.}
\label{fig:ladder}
\end{figure}

The question a scaling reviewer asks: does more data make the trunk
absorb the notebook? On the Paper-3 byte ladder (WikiText-103 slices;
same model recipe minus the positional table, varlen 64--256;
budget-matched arms), at 8/32/128/512\,MB
(Figure~\ref{fig:ladder}):

\begin{table}[h]
\centering
\small
\begin{tabular}{lccc}
\toprule
rung & trunk bpb@256 & $+$nb bpb@256 & extra repeat gain @2048 ($8\times$) \\
\midrule
8\,MB & 2.2197 & 2.2166 & $\mathbf{+0.064}$ \\
32\,MB & 2.0122 & 2.0041 & $\mathbf{+0.110}$ \\
128\,MB & 1.9652 & 1.9596 & $\mathbf{+0.146}$ \\
512\,MB & 1.9650 & 1.9573 & $\mathbf{+0.156}$ (seeds 42/43: $+0.148$/$+0.163$) \\
\bottomrule
\end{tabular}
\caption{The ladder: the notebook's value rises monotonically with
data, at zero-to-negative overall cost on every rung.}
\label{tab:ladder}
\end{table}

The value of the notebook \textbf{rises monotonically with data} ---
$64\times$ data, $2.4\times$ gain --- at zero-to-negative overall cost
on every rung, and the 512\,MB point carries a two-seed error bar
($\pm 0.008$). The bare trunk's repeat gain is negative on every rung
($-0.04 \dots -0.05$): long-range exact recall does not emerge from
data scale in this family; it must be built in. (Sanity: the 32\,MB
trunk point, 2.01 bpb, reproduces the corresponding Paper-3 ladder
point~\cite{kathleen3}.) The parameter-matched attention twin, run on
the 512\,MB rung, posts a $+0.79$ extra repeat gain --- natural-text
repetition is its home game, as Section~5.4 scoped --- while remaining
$O(L^2)$ and unevaluable at 4096 within the same memory budget; the
comparison the paper makes is exact recall at $O(L)$, and there the
twin scores zero (Section~3).

\section{Related work}

\textbf{State-space and gated-recurrent models} (S4~\cite{gu2022s4},
Mamba~\cite{gu2023mamba}, RWKV~\cite{peng2023rwkv},
Griffin~\cite{de2024griffin}): $O(L)$ trunks with fading state;
documented weakness at exact long-range recall; our trunk is
architecturally kin and inherits the diagnosis, which the notebook
addresses.

\textbf{Linear attention / fast weight programmers}
\cite{katharopoulos2020,schlag2021,yang2024deltanet}: the delta-rule
notebook of Round~1 is this family; the literature's
length-generalization failures match our Rounds~2--5, and the
content-key redesign is the departure point.

\textbf{Explicit memory modules}
(NTM/DNC~\cite{graves2014ntm,graves2016dnc}; Memorizing
Transformers~\cite{wu2022memorizing}; Titans~\cite{behrouz2025titans};
memory layers~\cite{berges2024memory}): learned read/write against a
separate store, typically with learned keys and/or attention-based
addressing. The notebook differs in its fixed, content-derived,
provably length-blind addressing and its 25K-parameter budget.

\textbf{Holographic reduced representations / VSA}
\cite{plate1995hrr,kanerva2009}: the binding/bundling algebra and
unitary keys are classical; the contributions here are the learned
\emph{when-to-write} under an LM loss, the self-gating raw read,
write-triggered forgetting, and the length-invariance measurements in
a trained end-to-end model.

\textbf{Length generalization in transformers} (RoPE
extensions~\cite{su2024rope}, YaRN~\cite{peng2024yarn},
ALiBi~\cite{press2022alibi}): an entire engineering literature works
around the attention length cliff we reproduce at small scale; the
notebook sidesteps rather than patches it, at $O(L)$.

\textbf{Retrieval-augmented decoding} (Paper 3,
\S5.2~\cite{kathleen3}): non-parametric phrase memory at \emph{decode}
time; the notebook is the \emph{train}-time, in-context counterpart.
The provenance finding there (own-corpus only) and the habitat finding
here (bytes only) are the same lesson at two levels: memory helps
where its answers are exact.

\section{Limitations}

\begin{itemize}
\item \textbf{Scale.} All results are 0.2--3.4M parameters,
  $\leq 512$\,MB data, single GPU. The ladder's monotonic trend is
  four points with a two-seed error bar at the top; the
  compute-optimal regime beyond ${\sim}0.5$M parameters remains
  unexplored.
\item \textbf{Salience without structure.} Un-marked needle detection
  is half-solved (53--64\%); the wide local gate alone does not solve
  it (4\%), and a hybrid local$+$context gate is future work. On
  natural text this matters less (the gate finds content words), but
  structured domains without markers will need it.
\item \textbf{Attention comparison scope.} The attention twin is one
  causal head on the same trunk, budget-matched --- the honest small
  mirror, not a tuned modern transformer with rotary extrapolation
  tricks.
\item \textbf{Forgetting trades gain for safety.} The eraser costs
  some near-distance gain; the algebra (unitary keys $\Rightarrow$ no
  targeted erase) says this is inherent to the superposition design,
  but alternative store geometries (e.g., slot-based) could reopen it.
\item \textbf{Word-level absorption} is reported as a boundary of the
  method, not resolved.
\end{itemize}

\section{Conclusion and the road ahead}

Paper 3 ended by naming the fading state's weakness; this paper
removes it without betraying the series' constraints. A 25K-parameter
notebook --- fixed holographic content keys, a myopically local
learned write gate, a read that gates itself, and forgetting tied to
writing rather than to time --- gives an attention-free byte model
one-shot recall that survives $4$--$8\times$ the training length, pays
on real text precisely where the recurrent state fades, costs nothing
anywhere, and \emph{appreciates} with data scale. Because the store is
algebraic, the model also does two things gradient-trained weights
cannot: it forgets on command with a guarantee (one subtraction, one
fact, chance-level erasure, no collateral) and it shows its sources
(100\% per-token provenance by counterfactual erasure). The route
mattered as much as the destination: three separate length leaks, none
in the memory itself, each found by oracle and counterfactual rather
than by search.

The road ahead follows the series' arc. (1)~\emph{Scale}: a
compute-optimal pass over the two-layer recipe.
(2)~\emph{Salience}: hybrid gating for structure-free domains.
(3)~\emph{Signals}: the notebook's habitat is raw streams --- the
natural next flagship is audio and sensor data, where ``remember the
signature you saw once, however far back, on-device'' is not a
benchmark but the product. Kathleen reads, writes, and now remembers;
next she hears.

\paragraph{Disclosure.} The memory mechanism described in this paper
(content-keyed holographic superposition with locally learned write
gates, raw self-gating readout, and write-triggered forgetting) is the
subject of U.S. Provisional Patent Application No.~64/140,260, filed
August~25, 2026.

\appendix

\section{Recipe card (exact hyperparameters)}

Trunk: $d = 96$ (needle rig) / 128 (ladder), 2--3 blocks, multi-scale
reverb (fast 0.50--0.90, med 0.90--0.99, slow 0.95--0.9995), causal
conv $k=7$, FFN $\times 2$, dropout 0.1 (text), no positional table.
Notebook: $\hrr = 2048$, content keys $\mathcal{N}(0, I)$
$\ell_2$-normalized (buffer), role keys unitary (7 for bytes, 3 for
words), write gate: $\mathrm{Embedding}(V, 32) \to$ causal
$\mathrm{Conv1d}(32, 32, k{=}16) \to \mathrm{GELU} \to
\mathrm{Linear}(32, 1)$, bias init $+2$; scale init 2000; forget
$g = \sigma(g_{\mathrm{logit}})\cdot 0.2$, $g_{\mathrm{logit}}$ init
$-2$; chunked scan, chunk 128--256, keep clamped $\geq 0.75$.
Training: AdamW lr $3\mathrm{e}{-3}$ (rig) / $2\mathrm{e}{-3}$
(ladder), wd 0.01, clip 1.0, batch 24--64, varlen per batch. Eval:
teacher-forced value-byte accuracy (needle); bits/byte with repeat
masks (text). Seeds 42/43; all construction after seeding (see
R3--R5 harness note).

\section{Full experiment registry}

\begin{table}[h]
\centering
\scriptsize
\begin{tabular}{clll}
\toprule
\# & experiment & verdict (pre-registered) & key numbers \\
\midrule
1 & NEEDLE R1 --- delta notebook, state gates & NOTEBOOK & 100\% train / 97\% ZS (s42) \\
2 & NEEDLE R2 --- multi-needle, no-mark & BOTH FAIL & 3--4\% all hard arms \\
3 & NEEDLE R3 --- budget vs curriculum & (confounded; superseded) & harness RNG-order bug found \\
4 & NEEDLE R4 --- replication/seed sweep & ROBUST SINGLE & click timing seed-dependent \\
5 & NEEDLE R5 --- multi-needle budget/capacity & SLOW CLICK & click at 2--6K steps; ZS 0\% \\
6 & NEEDLE R6 --- fixed content keys & (threshold artifact) & instant learning; nomark 53--64\%; ZS 0 \\
7 & NEEDLE R7 --- gates $\times$ capacity $2{\times}2$ & (capacity solved) & 2048: 96--97\%; local gates dead; ZS ${\sim}0$ \\
8 & NEEDLE R8 --- oracle gates + autopsy & MEMORY LEAK (superseded) & oracle ZS 0; $\beta$ sag $0.49 \to 0.11$ \\
9 & local diagnostics --- untrained memory & (length-proof) & 90.3/89.6\% @ 512/2048, zero training \\
10 & NEEDLE R9 --- raw read + wide local gate & GATE SHORTFALL $\to$ drift & local gates alive 85--86\%; ZS 0.5\% \\
11 & NEEDLE R10 --- varlen training & \textbf{CURED} & 81\% ZS learned gates $\times$2 seeds \\
12 & TEXT 1 --- repeat probe WT-2 & \textbf{REAL LICK} & extra $+0.15$--$0.27$; $\beta$ 0.28 vs 0.19 \\
13 & TEXT 2 --- distance bins + attn twin & FADES (scoped) & attn $+0.83$--$1.04$ on text; junk cost @4096 \\
14 & TEXT 3 --- forgetting & CLEAN INK & first-cost $+0.33 \to -0.004$ \\
15 & UNION --- word-level composer & SWALLOWED (structural) & extra $\leq +0.13$; $\beta$ flat \\
16 & LADDER --- 8/32/128\,MB & \textbf{CLIMBS} & $+0.064 \to +0.110 \to +0.146$ \\
17 & LADDER-512 --- big rung, twin, 2 seeds & \textbf{CLIMBS TO 512} & $+0.156$ mean; attn $+0.79$ on text \\
18 & UNLEARN --- erase one fact by subtraction & \textbf{UNLEARNED} & erased 4.3/3.1\% (chance 3.8); retained 84\% \\
19 & ATTRIB --- component + fact provenance & \textbf{ATTRIBUTED} & provenance 100\%/100\%; filler $\Delta \approx 0$ \\
\bottomrule
\end{tabular}
\caption{Every experiment in the program, with its pre-registered
verdict. JSON reports accompany the scripts.}
\label{tab:registry}
\end{table}

\end{document}